\documentclass[10pt,twocolumn,letterpaper]{article}

\usepackage{cvpr}
\usepackage{times}
\usepackage{epsfig}
\usepackage{graphicx}
\usepackage{amsmath}
\usepackage{amssymb}
\usepackage{color}
\usepackage{times}
\usepackage{epsfig}
\usepackage{algorithm}
\usepackage{algorithmicx}
\usepackage{algpseudocode}
\usepackage{graphics}
\usepackage{threeparttable}
\usepackage{color}
\usepackage[normalem]{ulem}
\usepackage{multirow}
\usepackage{float}
\usepackage{amsfonts}
\usepackage{bm}
\usepackage{subfig}
\usepackage{array}
\usepackage[table]{xcolor}
\usepackage{colortbl}
\usepackage{pifont}
\usepackage{cite}
\usepackage{diagbox}
\usepackage{rotating}
\usepackage{booktabs}
\usepackage{overpic}
\usepackage{textcomp}
\usepackage{contour}
\usepackage{enumitem}
\usepackage{ulem}

\usepackage[pagebackref=true,breaklinks=true,letterpaper=true,colorlinks,bookmarks=false]{hyperref}
\cvprfinalcopy 

\def\cvprPaperID{****} 

\ifcvprfinal\fi
\begin{document}

\title{Taylor saves for later: disentanglement  for video prediction using  taylor representation}
\author{Ting Pan $^{1,b}$\quad  Zhuqing Jiang$^{1,2,a,b}$\quad  Jianan Han$^1$ \quad Shiping Wen$ ^3 $\quad Aidong Men$^{1,a}$\quad Haiying Wang$^1$\\
	\small{$^a$} \small Corresponding author\\
	\small{$ ^b $}\small The first two authors contribute equally to this work\\
	\small{$^1$} \small School of Artificial Intelligence, Beijing University of Posts and Telecommunications,
Beijing, China \\
	\small{$ ^2 $}\small Beijing Key Laboratory of Network System and Network Culture, Beijing University of
Posts and Telecommunications, Beijing, China\\
	\small{$ ^3 $}\small Australian AI Institute, Faculty of Engineering and Information Technology, University of Technology Sydney, Australia\\
	{\tt\small $\{$panting,jiangzhuqing$\}$@bupt.edu.cn  \quad 
}
}

\maketitle
\thispagestyle{empty}

\vspace{-8pt}
\begin{abstract}
Video prediction is a challenging task with wide application prospects in meteorology and robot systems. 
Existing works fail to trade off short-term and long-term prediction performances and extract robust latent dynamics laws in video frames.
We propose a two-branch seq-to-seq deep model to disentangle the Taylor feature and the residual feature in video frames by a novel recurrent prediction module (TaylorCell) and residual module.
TaylorCell can expand the video frames' high-dimensional features into the finite Taylor series to describe the latent laws. 
In TaylorCell, we propose the Taylor prediction unit (TPU) and the memory correction unit (MCU). TPU employs the first input frame's derivative information to predict the future frames, avoiding error accumulation. MCU distills all past frames' information to correct the predicted Taylor feature from TPU.
Correspondingly, the residual module extracts the residual feature complementary to the Taylor feature.
On three generalist datasets (Moving MNIST, TaxiBJ, Human 3.6), our model outperforms or reaches state-of-the-art models, and ablation experiments demonstrate the effectiveness of our model in long-term prediction.

\end{abstract}

\section{Introduction}

Video prediction refers to predicting future multiple frames according to given frames, regarded as an intermediate step between the original video data and the decision-making system, which extracts potential evolution patterns from the original video data. It has wide application prospects in the fields of Meteorology \cite{Shi2015ConvolutionalLN,Shi2017DeepLF}, transportation \cite{EDACNN2020,ASADI2020105963,Hu2020ProbabilisticFP,Castrejn2019ImprovedCV,Jin2020ExploringSM,Wang2019Eidetic3L}, robot system \cite{Finn2017DeepVF,Xu2020VideoPV,Finn2016UnsupervisedLF} and anomaly detection \cite{Liu2018FutureFP}.

\begin{figure}[t]
    \centering
      \includegraphics[width=1.0\linewidth]{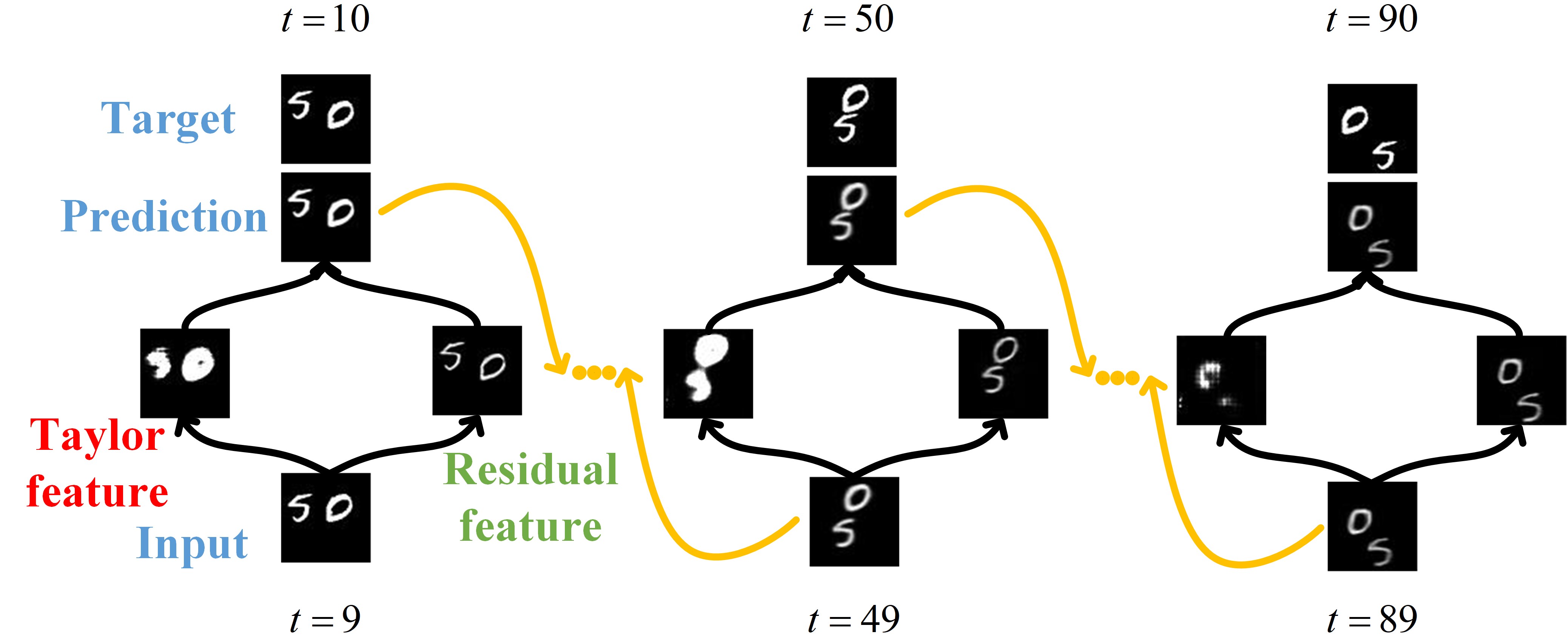}
	  \caption{Visualization of Taylor feature and residual feature in our model. 
	  	With the disentanglement of our model, the residual feature learns clear content while the Taylor feature learns salient edges and corners, which tend to blur in the long-term prediction.}
	  \label{fig:introduce} 
\end{figure}
Models representing temporal dependence are often applied to predict video frames. Mainstream video prediction models can mostly be categorized into three frameworks: (1) extensions of recurrent neural networks \cite{Shi2015ConvolutionalLN,Lange2020AttentionAC,Wang2017PredRNNRN,Wang2019MemoryIM,Wang2018PredRNNTA} or 3D convolution \cite{Wang2019Eidetic3L,Vondrick2016GeneratingVW,Tulyakov2018MoCoGANDM}; (2) conditioning the prediction on proxy objects \cite{Wang2018VideotoVideoS,Villegas2017LearningTG}; (3) specific architectures based on factorized the prediction space \cite{Vondrick2016GeneratingVW,Sreekar2020MutualIB,Guen2020DisentanglingPD,Dona2020PDEDrivenSD,Hsieh2018LearningTD}.

These frameworks hold advantages and inferiors of themself. (1) the first framework attempts to design a new spatiotemporal module that enables better spatial relationships and temporal dependence and then stacks multiple such modules to form the final model.
However, These models are usually challenged by more parameters and complex training.
(2) The second framework narrows the prediction space with the assistant of proxy objects that provide valuable information for the prediction task.
These models improve problems in the first framework but are burdened with extra data acquisition.
(3) The third framework factorizes the video into multiple variables to process each on a separate pathway.
These models design targeted modules to decompose the prediction task into more tractable problems and free from the chains of numerous parameters and supplementary data. 
Nevertheless, most disengagement models explicitly untangle video sequences in a way consistent with human intuition \cite{Vondrick2016GeneratingVW,Sreekar2020MutualIB,Hsieh2018LearningTD}, such as foreground and background.
These methods contribute to precise short-term prediction yet fail to maintain their performance in a long-term setting.
PhyDNet \cite{Guen2020DisentanglingPD} separates physical dynamics and unknown factors in video and uncovers hidden partial differential equations (PDE) models from observed complex dynamic video data. 
Although boosting long-term prediction performance, it suffers from error accumulation.
PDSD \cite{Dona2020PDEDrivenSD} disentangles spatial and temporal representations for prediction. 
It works well for long-term prediction and content swap, whereas relatively blurrily predicts the next frames.
On the whole, current disentanglement models for video prediction fail to trade off short-term and long-term prediction performance.
\begin{figure*}[t]
	\centering
	\centerline{\includegraphics[width=\linewidth]{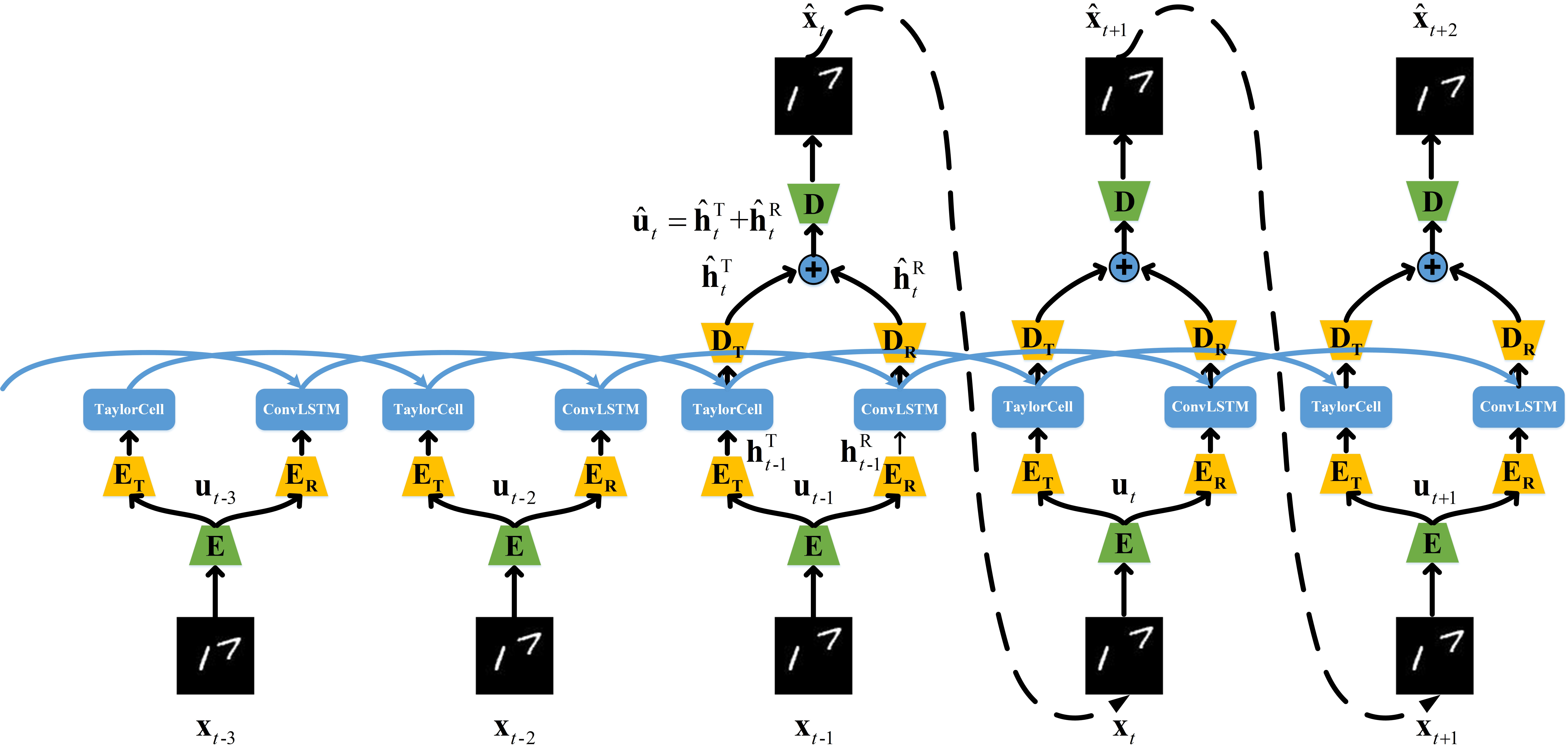}}
	\caption{The framework of our method. TaylorNet forms a two-branch seq-to-seq architecture for video prediction. 
		The input video frames are first mapped to the latent space $ U $, then are linearly disentangled as the Taylor feature $ \mathbf{h}_{t-1}^{\mathrm{T}} $ and the residual feature $ \mathbf{h}_{t-1}^{\mathrm{R}} $ by $ \mathbf{E}_{\mathbf{T}} $ and $ \mathbf{E}_{\mathbf{R}} $. 
		TaylorCell (section \ref{TaylorCell}) and ConvLSTM predict the Taylor feature $ \hat{\mathbf{h}}_{t}^{\mathrm{T}} $ and the residual feature  $ \hat{\mathbf{h}}_{t}^{\mathrm{R}} $ of the next time stamp. 
		Finally, $ \hat{\mathbf{h}}_{t}^{\mathrm{T}} $ and  $ \hat{\mathbf{h}}_{t}^{\mathrm{R}} $ are remapped to the latent space and are summed before decoding to generate the future frame $ \hat{\mathbf{u}}_{t} $.}
	\label{fig:TaylorNet}
\end{figure*}

Regarding the above problems, an algorithm that considers modeling both short-term and long-term dynamical systems is requested.
Taylor series that predict the value of other points by the finite derivatives of a certain point satisfies our needs. 
Therefore, in this work, we propose TaylorNet, as shown in Fig \ref{fig:TaylorNet}, a two-branch seq-to-seq deep model assisted by the Taylor series priors, which accurately predicts the future frames and outperforms the state-of-the-art model in both short-term and long-term prediction.
Furthermore, we propose a novel prediction module. 
It expands the first input frame and then constructs an approximate polynomial to form complex dynamics evolution in video.
The visualizations of the Taylor feature and the residual feature are shown in Fig \ref{fig:introduce}. 
The Taylor feature contains more information about edges and corners, regarded as effective complements to the residual feature. 
Our contributions to video prediction can be summarized as follows:

\begin{itemize}
	\item{We propose a novel principle for feature separation. Utilizing a two-branch network design, we present a new architecture for video prediction.}
	\item{Base on the principle, a novel recurrent prediction module (TaylorCell) is integrated into the two-branch model, which contains Taylor prediction unit (TPU) and memory correction unit (MCU). TPU only employs finite derivatives of the first input frame to predict the future frames for avoiding error accumulation; MCU corrects the predicted Taylor feature from TPU by distilling information of all past frames through the gate mechanism.
	} 
	\item{Experiments demonstrate that TaylorNet outperforms or reaches state-of-the-art models on three generalist datasets. As far as we know, this is the first video prediction model built by the Taylor series.}
\end{itemize}

The rest of this paper is organized as follows. 
In section \ref{Methodology}, the proposed method is described in detail.
Section \ref{Experiment}  presents extensive experiments.
Section \ref{Conclusion} is the conclusion.

\section{Related Works}
\label{Related_work}

We review related models of video prediction in recent years.


\textbf{Extensions of RNN or  3D-Conv}	The last few years have witnessed a significant interest in predicting future frames by data-driven deep neural networks. 
Shi  \etal \cite{Shi2015ConvolutionalLN} extended vanilla LSTM to ConvLSTM by integrating convolutions into recurrent state transitions, capturing spatial information for precipitation prediction.
PredRNN \cite{Wang2017PredRNNRN} allowed memory states to zigzag in two directions: across stacked RNN layers vertically and through all RNN states horizontally.
Causal LSTM \cite{Wang2018PredRNNTA} increased the transition depth between adjacent states by re-organizing the spatial and temporal memories in a cascaded mechanism.
E3D-LSTM \cite{Wang2019Eidetic3L} integrated 3D convolutions into RNNs, which enable the memory cell to store better short-term features and made the present memory state interact with its historical records to learn long-term relations.
Lange \etal \cite{Lange2020AttentionAC} introduced the self-attention mechanism into ConvLSTM to extract spatial features with both global and local dependencies.
The above-mentioned recurrent models mainly employed stacked ConvRNNs to predict future frames, which stored better spatiotemporal features, but consumed considerable GPU memory and computational power.

\textbf{Conditioning the prediction on proxy objects}	Some models \cite{Terwilliger2019RecurrentFS,Reda2018SDCNetVP} utilized the FlowNet architecture \cite{Ilg2017FlowNet2E,Dosovitskiy_2015_ICCV} with pre-trained weights from FlowNet2 \cite{Ilg2017FlowNet2E} for video prediction.
Villegas \etal \cite{Villegas2017LearningTG} estimated high-level structure in the input frames, \textit{e.g.} the pose landmarks, then predicted how that structure evolved in the future.
Wang \etal \cite{Wang2018VideotoVideoS} extracted the segmentation masks from the input frames to predict that in next time stamp, then used a video-to-video synthesis approach to convert the predicted segmentation masks to a future video. 
Ye \etal \cite{Ye2019CompositionalVP} learned to predict the future locations and appearance of the entities in the scene by (known or detected) locations of the entities present to compose a future frame prediction.  
These models required supplementary data or a specific pre-trained functional network, \textit{e.g.} the optical flow network.
 
\textbf{Factorized the prediction space}	Some models \cite{Vondrick2016GeneratingVW, Sreekar2020MutualIB,Guen2020DisentanglingPD,Dona2020PDEDrivenSD} disentangled video frames into multiple variables, \textit{e.g.} foreground and background, content and low dimensional pose feature, physical dynamics and unknown factors, and spatial and dynamic feature. 
However, these models failed to trade off short-term and long-term prediction performance or suffer from error accumulation.
   
\textbf{Stochastic prediction}	Models \cite{Franceschi2020StochasticLR,Mohammad2018SV2P,Denton2018SVG,Castrejn2019ImprovedCV,Ruben2019HFVP,Richard2018SAVP} applied VAE-based, GAN-based or latent variable-based approaches to account for the inherent stochasticity in video sequences by building distributions over possible futures, yet beyond our study's scope.

\textbf{Taylor series}	An increasing number of works combining neural networks and Taylor disengagement for deep neural network interpretability have been produced for the last few years. 
Montavon \etal \cite{Montavon2017ExplainingNC} considered the problem of explaining classification decisions of a deep network in terms of input variables and backpropagated the explanations from the output to the input layer based on deep Taylor disengagement.
The model viewed each neuron of a deep network as a function that can be expanded and decomposed on its input variables.
Based on the above work, Hiley \etal \cite{Hiley2019DiscriminatingSA} produced a naive representation of both the spatial and temporal relevance of a frame as two separate objects.
The model separated the motion relevance by downweighting relevance in the original explanation by the spatial relevance reconstructed from each frame’s explanation.
Although these works leveraged Taylor expansions, they combined the Taylor decomposition with backpropagation rules and were inapplicable to video prediction.

\section{Methodology}
\label{Methodology}
This section develops the TaylorNet, a two-branch seq-to-seq model for video prediction.
We introduce TaylorCell to separate the Taylor representations from other factors in video.
TaylorCell includes the Taylor prediction unit (TPU, section \ref{TPU}) and the memory correction unit (MCU, section \ref{MCU}). 

\subsection{Problem Statement}
\label{Problem Statement}
We define $ \mathbf{X}_{i}\in\mathbf{Q}^{w \times h \times c} $ as the $ i $-th frame in the input video frames $ \mathbf{X}_{input}=\left(\mathbf{x}_{0},\mathbf{x}_{1},\ldots, \mathbf{x}_{t-1}\right) $ with $ t $ frames,where $ w $, $ h $, and $ c $ denote width, height, and number of channels, respectively. The target of video prediction is to predict the next frames $ \mathbf{X}_{output}=\left(\mathbf{x}_{t},\mathbf{x}_{t+1},\ldots,\mathbf{x}_{t+N}\right) $ from  $ \mathbf{X}_{input} $. We define $ i $-th frame $ \mathbf{x}_{i}=\mathbf{x}\left(i, x, y\right) $ ,where $ x $ and $ y $ represent spatial coordinates. 
We assume that there exists a latent space $ U $ in which the Taylor representations and the residual factors of the video frames can be linearly separated.  
Based on the assumption, $ \mathbf{u}_{i} \in U $ denotes the latent representation of the $ i $-th frame, and enables to decompose as $  \mathbf{u}_{i}=\mathbf{h}_{i}^{\mathrm{T}}+\mathbf{h}_{i}^{\mathrm{R}} $, where $\mathbf{h}_{i}^{\mathrm{T}}  $ and $ \mathbf{h}_{i}^{\mathrm{R}} $ denote the Taylor feature and the residual feature, respectively.

\subsection{Network Architecture}
\label{Network Architecture}
We first outline the general pipeline of our method. Our TaylorNet consists of two subnetworks, a recurrent predictor based on the Taylor series priors and a generic recurrent neural network for residual factors. 
The architecture of the proposed TaylorNet is illustrated in Fig \ref{fig:TaylorNet}.
The left branch models the Taylor feature, moreover the right branch forms the residual components that TaylorCell fails to extract.

Formally, the input ($ t-1$)-th frame is first encoded as the latent features $  \mathbf{u}_{t-1} $, then mapped to Taylor feature space and residual feature space by $ \mathbf{E}_{\mathbf{T}} $ and $ \mathbf{E}_{\mathbf{R}} $.
 The output $ \mathbf{h}_{t-1}^{\mathrm{T}} $ and $ \mathbf{h}_{t-1}^{\mathrm{R}} $ stand for the Taylor feature and the residual feature, respectively.

We propose TaylorCell, which combines Taylor series priors to predict the Taylor feature of the next time stamp in the left branch. There are 2 inputs: the hidden states from the previous time stamp and Taylor feature $ \mathbf{h}_{t-1}^{\mathrm{T}} $ from the current input frame. 
To integrate with the right branch, we remap the predicted Taylor feature to the latent space $ U $ by the decoder $ \mathbf{D}_{\mathbf{T}} $. 
For the convenience of description, we define $ \hat{\mathbf{h}}_{t}^{\mathrm{T}} $ as the $ t$-th predicted Taylor feature.

Furthermore, we apply a data-driven vanilla ConvLSTM without any priors to predict the residual feature of the next time stamp in the right branch. 
There are 2 inputs: the hidden states from the previous time stamp and residual feature $ \mathbf{h}_{t-1}^{\mathrm{R}} $ from the current input frame.
In the same manner, we remap the residual Taylor feature to the latent space $ U $ by the decoder $ \mathbf{D}_{\mathbf{R}} $, where $ \hat{\mathbf{h}}_{t}^{\mathrm{R}} $ denotes the $ t$-th predicted residual feature.
This subnetwork aims to extract unknown factors as effective complements to the Taylor feature.
In other words, the Taylor feature assists the prediction of the residual feature, as shown in Fig \ref{fig:introduce}, which is detailed in \textbf{Orders of the Taylor Series} of section \ref{Ablation Study}.

With the predicted Taylor feature $ \hat{\mathbf{h}}_{t}^{\mathrm{T}} $  and the predicted residual feature $ \hat{\mathbf{h}}_{t}^{\mathrm{R}} $ , the output predicted latent feature is:
\begin{equation}
\hat{\mathbf{u}}_{t}=\hat{\mathbf{h}}_{t}^{\mathrm{T}}+\hat{\mathbf{h}}_{t}^{\mathrm{R}} 
\end{equation}
Finally, $ \hat{\mathbf{u}}_{t} $ is mapped to the low-dimensional space by the decoder $ \mathbf{D} $  to generate the $ t $-th predicted frame $ \hat{\mathbf{x}}_{i} $.

\begin{figure}[t]
	\centering
	\includegraphics[width=1.0\linewidth]{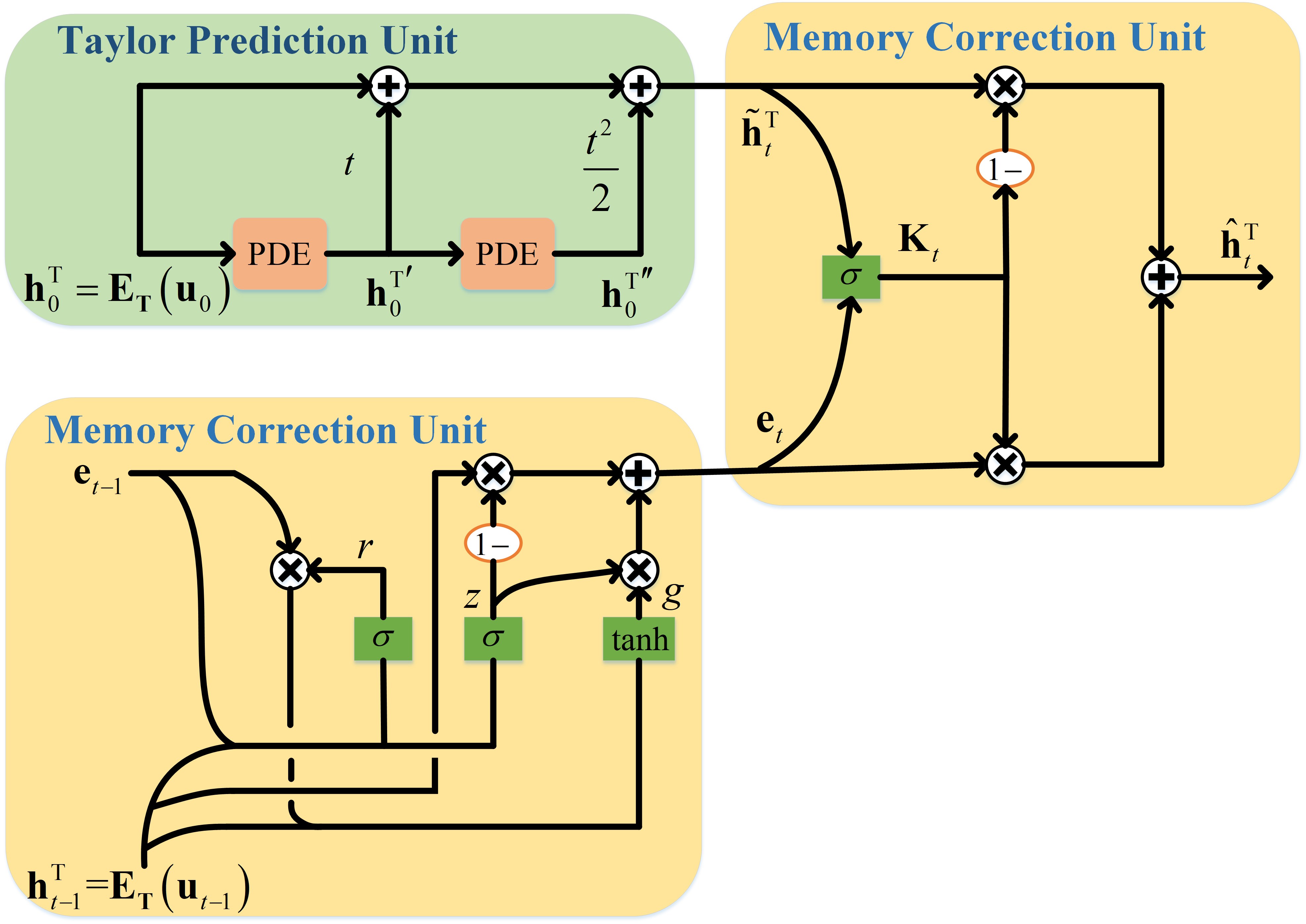}
	\caption{TaylorCell introduces two units: (1) the green block demonstrates TPU, which predicts the Taylor feature with the first frame's spatial information and the current frame's time parameters; (2) the yellow block represents MCU, which extracts all past frames' information through the gate mechanism to correct the predicted Taylor feature from TPU.}
	\label{fig:TaylorCell} 
\end{figure}

\subsection{TaylorCell}
\label{TaylorCell}

TaylorCell is a novel recurrent prediction module, as shown in Fig \ref{fig:TaylorCell}, which is motivated by the Taylor series and Kalman filter \cite{10.1115/1.3662552}. 
It's used to model latent laws in video sequences for memorizing and distilling valuable information, which the finite Taylor series have perceived.
TaylorCell includes two units: Taylor prediction unit, which predicts the Taylor feature with spatial information of the first frame and time parameters of the current frame; Memory correction unit, which extracts information of all past frames through the gate mechanism to correct the predicted Taylor feature from the Taylor prediction unit.

\subsubsection{Taylor Prediction Unit}
\label{TPU}

The green block of Fig \ref{fig:TaylorCell} illustrates the framework of TPU. 
To avoid error accumulation in long-term prediction, we solely expand the first input frame into the finite Taylor series to conduct the mathematical calculation instead of forward Euler in time.
We define Taylor inferred feature as $ \tilde{\mathbf{h}}_{t}^{\mathrm{T}} $ at the time step of $ t $. Therefore, the function of TPU can be expressed as:
\begin{equation}
\tilde{\mathbf{h}}_{t}^{\mathrm{T}}=\underbrace{\mathbf{h}_{0}^{\mathrm{T}}+t \mathbf{h}_{0}^{\mathrm{T} \prime}+\frac{t^{2}}{2} \mathbf{h}_{0}^{\mathrm{T} \prime \prime}}_{\xi=3}+\cdots
\end{equation}
where $ \mathbf{h}_{0}^{\mathrm{T}} $, $ \mathbf{h}_{0}^{\mathrm{T} \prime} $ and $ \mathbf{h}_{0}^{\mathrm{T} \prime \prime} $ denote the Taylor feature of the first input frame, as well as its first order and second order temporal derivative, respectively.
$ \mathbf{h}_{0}^{\mathrm{T} \prime} $ is generated by inputing $ \mathbf{h}_{0}^{\mathrm{T}} $  into the PDE model \cite{Long2019PDENet2L, Guen2020DisentanglingPD, pmlr-v80-long18a}, which can realize the time difference of the deep feature map. Moreover, we obtain $ \mathbf{h}_{0}^{\mathrm{T} \prime \prime} $ in the same way. 
In TPU, the expansion order $ \xi $ of the Taylor series is a superparameter, and we set $ \xi=3 $. More details are in \textbf{Orders of the Taylor Series} of section \ref{Ablation Study}.

In PDE model, we approximate the time difference through the finite space difference, as shown in Equ \ref{equ:partial differential equation family}. This partial differential equation family subsumes many classical physical models, such as thermodynamic equations, wave equations, and advection-diffusion equations.
\begin{equation}
\label{equ:partial differential equation family}
\mathbf{h}_{t}^{\mathrm{T} \prime}=\frac{\partial \mathbf{h}_{t}^{\mathrm{T}}}{\partial t}=\sum_{i, j : i+j \leq q} c_{i, j} \frac{\partial^{i+j} \mathbf{h}_{t}^{\mathrm{T}}}{\partial x^{i} \partial y^{j}}
\end{equation}
where $ x $ and $ y $ represent spatial coordinates, and $ c_{i, j} $ demonstrates coefficients of linear combination, generated by $ 1\times1 $ filters.

The finite space difference is realized by convolutions with trainable filters.
For a $ k \times k $ filter $ \mathbf{w} $, define its moment matrix as:
\begin{equation}
\mathbf{M}(\mathbf{w})_{i, j}=\frac{1}{i ! j !} \sum_{u,v=-\frac{k-1}{2}}^{\frac{k-1}{2}} u^{i} v^{j} \mathbf{w}[u, v] 
\end{equation}
where $ u$ and $ v $ denote spatial coordinates in filter $ \mathbf{w} $, $ i $ and $ j $ denote spatial coordinates in moment matrix $ \mathbf{M}(\mathbf{w}) $, and $ i, j=0, \ldots, k-1 $. When feature map $ \mathbf{h} $ is convoluted with $ \mathbf{w} $, we expand $ \mathbf{h} $ to the Taylor series:
\begin{equation}
\begin{aligned}
\mathbf{w} \circledast \mathbf{h} = \sum_{i, j=1}^{k-1} \mathbf{M}(\mathbf{w})_{i, j} \delta x^{i} \delta y^{j} \frac{\partial^{i+j} \mathbf{h}}{\partial x^{i} \partial y^{j}}(x, y) +o\left(\bullet\right)
\end{aligned}
\end{equation}
where $ o\left(\bullet\right) $ denotes an infinitesimal. If we impose the following constrains on $ \mathbf{M}(\mathbf{w}) $:
\begin{equation}
\mathbf{M}(\mathbf{w})=\left(\begin{array}{lll}
0 & 0 & 0 \\
1 & 0 & 0 \\
0 & 0 & 0
\end{array}\right)
\end{equation}
$ \frac{\partial \mathbf{h}}{\partial x} $ can be approximated by $ \mathbf{w} \circledast \mathbf{h} $, where $ \mathbf{w} $ is a $ 3\times3 $ filter. 

Therefore, if multi-channels $ \mathbf{M}(\mathbf{w}) $ equals 1 at position $ \left( i,j\right)  $ and 0 elsewhere, each channel of convolution $ \mathbf{w} \circledast \mathbf{h} $ is a different-order space difference feature map. 
Then all channels linearly combine through $ 1\times1 $ filters to approximate the time difference feature map.
To impose constraints on $ \mathbf{M}(\mathbf{w}) $, a moment loss is applied:
\begin{equation}
\mathrm{L}_{\text {moment}}=\sum_{i,j \leq k}\left\|\mathbf{M}(\mathbf{w})_{i, j}-\Delta_{i, j}\right\|
\end{equation}
where $ \Delta_{i, j} $ denotes a matrix with 1 at position $ \left( i,j\right)  $ and 0 elsewhere. When moment loss is slight, the function of  $ \mathbf{w} $  approximates $ \frac{\partial^{i+j}}{\partial x^{i} \partial y^{j}}(\cdot) $.

\subsubsection{Memory Correction Unit}
\label{MCU}

There exist three phenomena:
(1) factors of a long video sequence may change at some timestamps;
(2) the Taylor series is limited to the convergence domain;
(3) finite Taylor expansions fail to approximate the complex dynamic system.
Hence, it's inaccurate to employ the Taylor inferred feature as the predicted Taylor feature at the next time stamp.

Inspired by the Kalman filter, MCU distills Taylor information of input frames to correct the Taylor inferred feature,  as shown in the yellow block of Fig \ref{fig:TaylorCell}. Particularly, in our model, Taylor information of all past frames is merged instead of sole the previous frame.
The $ t $-th hidden states $ \mathbf{e}_{t} $ in MCU is defined as:
\begin{equation}
\begin{aligned}
\mathbf{e}_{t}=\mathbf{z} \odot \mathbf{g}+(1-\mathbf{z}) \odot \mathbf{h}_{t-1}^{\mathrm{T}}
\end{aligned}
\end{equation}
where $ \mathbf{h}_{t-1}^{\mathrm{T}} $ and $ z $ denote Taylor feature of current input frame and update gate with value range (0,1), respectively. $ \mathbf{g} $ stands for updated memory state:
\begin{equation}
\mathbf{g}=\tanh \left(\mathbf{W}_{\mathrm{g}} \circledast \left(\mathbf{r e}_{t-1}, \mathbf{h}_{t-1}^{\mathrm{T}}\right)\right)
\end{equation}
where $ \mathbf{e}_{t-1} $ and $r$ denote hidden states of the previous time stamp in MCU and reset gate, respectively.

Finally, the predicted Taylor feature considers both the updated hidden states $ \mathbf{e}_{t} $ and the Taylor inferred feature $ \tilde{\mathbf{h}}_{t}^{\mathrm{T}} $ from TPU, defined as follows:
\begin{equation}
\label{equ:correct}
\hat{\mathbf{h}}_{t}^{\mathrm{T}}=\tilde{\mathbf{h}}_{t}^{\mathrm{T}}+\mathbf{K}_{t} \odot\left(\mathbf{e}_{t}-\tilde{\mathbf{h}}_{t}^{\mathrm{T}}\right)
\end{equation}
where $ \mathbf{K}_{t}\in{\left( 0,1\right) } $ demonstrates the correction parameters from inputing $ \tilde{\mathbf{h}}_{t}^{\mathrm{T}} $ and $ \mathbf{e}_{t} $ into $ 1\times1 $ filters. 
Note that if $ \mathbf{K}_{t}=0 $, the dynamics evolution follows the Taylor inferred feature; if $ \mathbf{K}_{t}=1 $ , the dynamics evolution is resetted and only driven by all past observations.

\subsection{Training}
\label{Training}

\begin{table}[t]
	\caption{Setting of learning rate, total epochs and batch size on three datasets}
	\centering
	\begin{tabular}{c|ccc}
		\hline
		\textbf{Datasets}     &\textbf{ LR} & \textbf{Epochs} & \textbf{Batch Size} \\ \hline
		Moving MNIST & 0.001         & 1000         & 16         \\
		TaxiBJ       & 0.0001        & 100          & 16         \\
		Human 3.6    & 0.0001        & 100          & 4          \\ \hline
	\end{tabular}
	\label{table:train}
\end{table}

Our unsupervised video prediction task aims to predict $ \mathbf{X}_{output}=\left(\mathbf{x}_{t},\mathbf{x}_{t+1},\ldots,\mathbf{x}_{t+N}\right) $ from input video frames $ \mathbf{X}_{input}=\left(\mathbf{x}_{0},\mathbf{x}_{1},\ldots, \mathbf{x}_{t-1}\right) $ absence of supplementary data.

\textbf{Mode}	In each epoch of training, we randomly choose whether to use the teaching mode. If teaching mode works, input frames of TaylorNet come from the original images; otherwise, input frames also come from the original images in $ 0 \sim \left( t-1\right)  $ time stamps, yet from the predicted video frames in $ t \sim \left( t+N\right)  $ time stamps. 

\textbf{Loss}	To generate the target video frames, we impose constrains on parameters of PDE model and penalize the difference between the original image and its reconstruction. So, full loss function $ \mathbf{L}(\mathbf{w}) $ linearly combines two partial losses:
\begin{equation}
\mathbf{L}(\mathbf{w})=\mathrm{L}_{\text {image }}(\mathbf{w})+\lambda \mathrm{L}_{\text {moment }}\left(\mathbf{w}_{_{\mathrm{PDE}}}\right)
\end{equation}
where $ \mathbf{w} $ represents all parameters in model, and $ \mathbf{w}_{_{\mathrm{PDE}}} $ denotes parameters in PDE model. We use the $ \mathbf{L}_{2} $ loss for the image reconstruction and the moment loss, and following \cite{Guen2020DisentanglingPD} we set $ \lambda=1 $. 

\textbf{Setting} Following the setting of \cite{Guen2020DisentanglingPD}, PDE model contains 49 filters ($7\times7 $) for space difference feature map and $ 1\times1 $  filters for linear combination.
The residual module is a 3-layers vanilla ConvLSTM.
We train with the adam optimizer on all datasets.
Learning rate, total epochs and batch size are set as Table \ref{table:train}.

\begin{table}[t]
	\caption{Quantitative comparison on Moving MNIST}
	\centering
	\begin{tabular}{c|c c c}
		\hline
		\textbf{Moving MNIST} && \textbf{10$ \to $10} \\
		\hline
		\textbf{Method} & \textbf{MSE} & \textbf{MAE} & \textbf{SSIM}  \\
		\hline
		ConvLSTM         &  103.3     & 182.9  &  0.707\\
		PredRNN          &  56.8     &   126.1  &    0.867\\
		Causal LSTM     & 46.5     & 106.8   &   0.898 \\
		MIM                 &  44.2      &  101.1  &  0.91 \\
		SA-ConvLSTM   &  43.9      & 94.7	 &    0.913 \\
		PDSD            & 42.9      & 105.6    &  0.878 \\
		E3D-LSTM       &  41.3      & 86.4    &   0.92\\
		DDPAE            &  38.9      & 90.7    &   0.922\\		
		CrevNet + ConvLSTM  &38.5   & -   & 0.928 \\
		PhyDNet            &  24.4     & 70.3    &   0.947\\
		PhyDNet-dual            &  23.5     & 68.6    &   0.948\\
		CrevNet + ST-LSTM & \color{blue}22.3     &-     &   \color{blue}0.949\\
		TaylorNet (ours)            & \color{red}22.2     &\color{red}65.2     &\color{red}0.952\\
		\hline
	\end{tabular}
	\label{table:mmnist}
\end{table}

\begin{table}[t]
	\caption{Quantitative comparison on TaxiBJ}
	\centering
	\begin{tabular}{c|ccc}
		\hline
		\textbf{TaxiBJ}      & \multicolumn{3}{c}{\textbf{4$ \to $ 4}} \\ \hline
		\textbf{Method}      & \textbf{MSE}$ \times\mathbf{100} $       & \textbf{MAE}       & \textbf{SSIM}       \\ \hline
		ConvLSTM    & 48.5          & 17.7      & 0.978      \\
		PredRNN     & 46.4          & 17.1      & 0.971      \\
		Causal LSTM & 44.8          & 16.9      & 0.977      \\
		E3D-LSTM    & 43.2          & 16.9      & 0.979      \\
		MIM         & 42.9          & 16.6      & 0.971      \\
		PhyDNet     & 41.9          & 16.2      & 0.982      \\
		SA-ConvLSTM & \color{red}39            &   -        & \color{red}0.984      \\
		TaylorNet (ours)        & \color{blue}40            & \color{red}16.2      & \color{blue}0.983     \\
		\hline
	\end{tabular}
	\label{table:taxibj}
\end{table}

\begin{table}[]
	\caption{Quantitative comparison on Human 3.6}
	\centering
	\begin{tabular}{c|ccc}
		\hline
		\textbf{Human 3.6}   & \multicolumn{3}{c}{\textbf{4 $ \to $ 4}} \\ \hline
		\textbf{Method}      & \textbf{MSE/10}      & \textbf{MAE/100}      & \textbf{SSIM}      \\ \hline
		ConvLSTM    & 50.4        & 18.9         & 0.776     \\
		PredRNN     & 48.4        & 18.9         & 0.781     \\
		E3D-LSTM    & 46.4        & 16.6         & 0.869     \\
		Causal LSTM & 45.8        & 17.2         & 0.851     \\
		MIM         & 42.9        & 17.8         & 0.79      \\
		PhyDNet     & \color{red}36.9        & \color{blue}16.2         & \color{red}0.901     \\
		TaylorNet (ours)        & \color{blue}37.4        & \color{red}16.1         & \color{red}0.901    \\
		\hline
	\end{tabular}
	\label{table:human}
\end{table}

\section{Experiment}
\label{Experiment}
In this section, we evaluate our TaylorNet framework quantitatively and
qualitatively on three generalist datasets: Moving MNIST \cite{Srivastava2015UnsupervisedLO}, TaxiBJ \cite{Zhang2017DeepSR}, and Human 3.6 \cite{Ionescu2014Human36MLS}.
We compare our model with the state-of-the-art video prediction methods based on standard metrics, \textit{i.e}., mean square error (MSE), mean absolute error (MAE), and structural similarity (SSIM).
The ablation experiments demonstrate the contribution of each module in our model.

\subsection{Datasets}
\label{Datasets}
\textbf{Moving MNIST} is a synthetic video dataset, which depicts two potentially overlapping digits moving with constant velocity and bouncing off the image edge. 
It's particularly challenging in the long-term prediction task.
Image size is $ 64\times64\times1 $.
We train TaylorNet in the same setting as \cite{Guen2020DisentanglingPD}, with 10 conditioning frames to predict 10 future frames, and evaluate our model by predicting either 10, 30 or 90 future frames.
Training sequences are generated on the fly and the test set of 10000 sequences is provided by \cite{Srivastava2015UnsupervisedLO}. 

\textbf{TaxiBJ} is collected from the chaotic real-world environment and contains traffic flow images collected consecutively from the GPS monitors of taxicabs in Beijing from 2013 to 2016. Each $ 32\times32\times2 $ image is a 2-channels heat map with leaving/entering traffic.
We use 4 known frames to predict the next 4 frames (traffic conditions for the next two hours).

\textbf{Human 3.6} contains human actions of 17 scenarios, including 3.6 million poses and corresponding images. Following the setting of \cite{Guen2020DisentanglingPD,Wang2019MemoryIM}, we use only the "walking" scenario with subjects S1, S5, S6, S7, S8 for training, and S9, S11 for testing. We predict 4 future frames of size $ 128 \times 128 \times 3 $ given 4 input frames.

\subsection{State-of-the-art comparison}
\label{State of the art comparison}
\begin{figure}[t]
	\centering
	\includegraphics[width=1.0\linewidth]{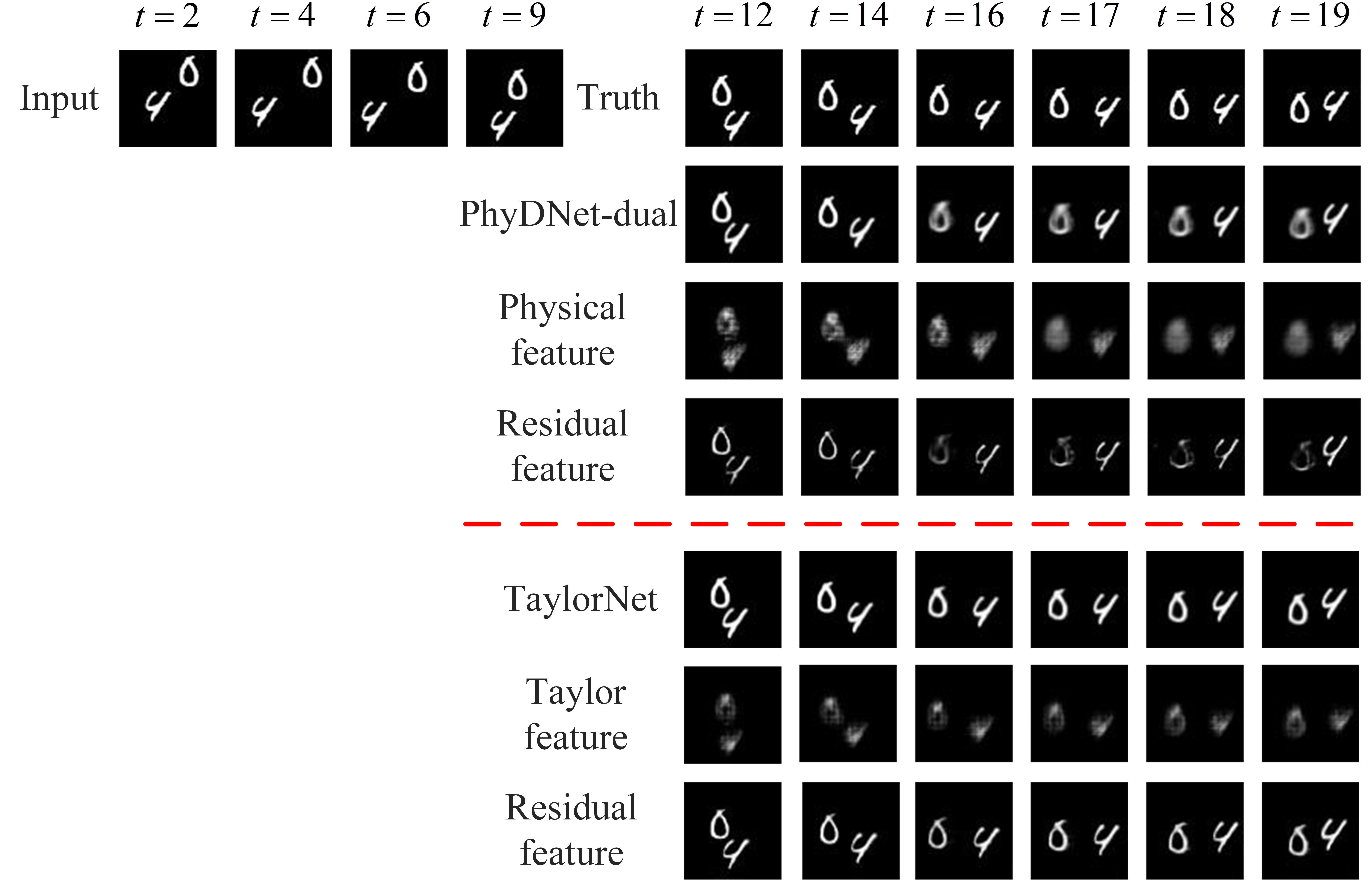}
	\caption{Qualitative comparison on Moving MNIST.}
	\label{fig:mmnist} 
\end{figure}

\begin{figure}[t]
	\centering
	\includegraphics[width=1.0\linewidth]{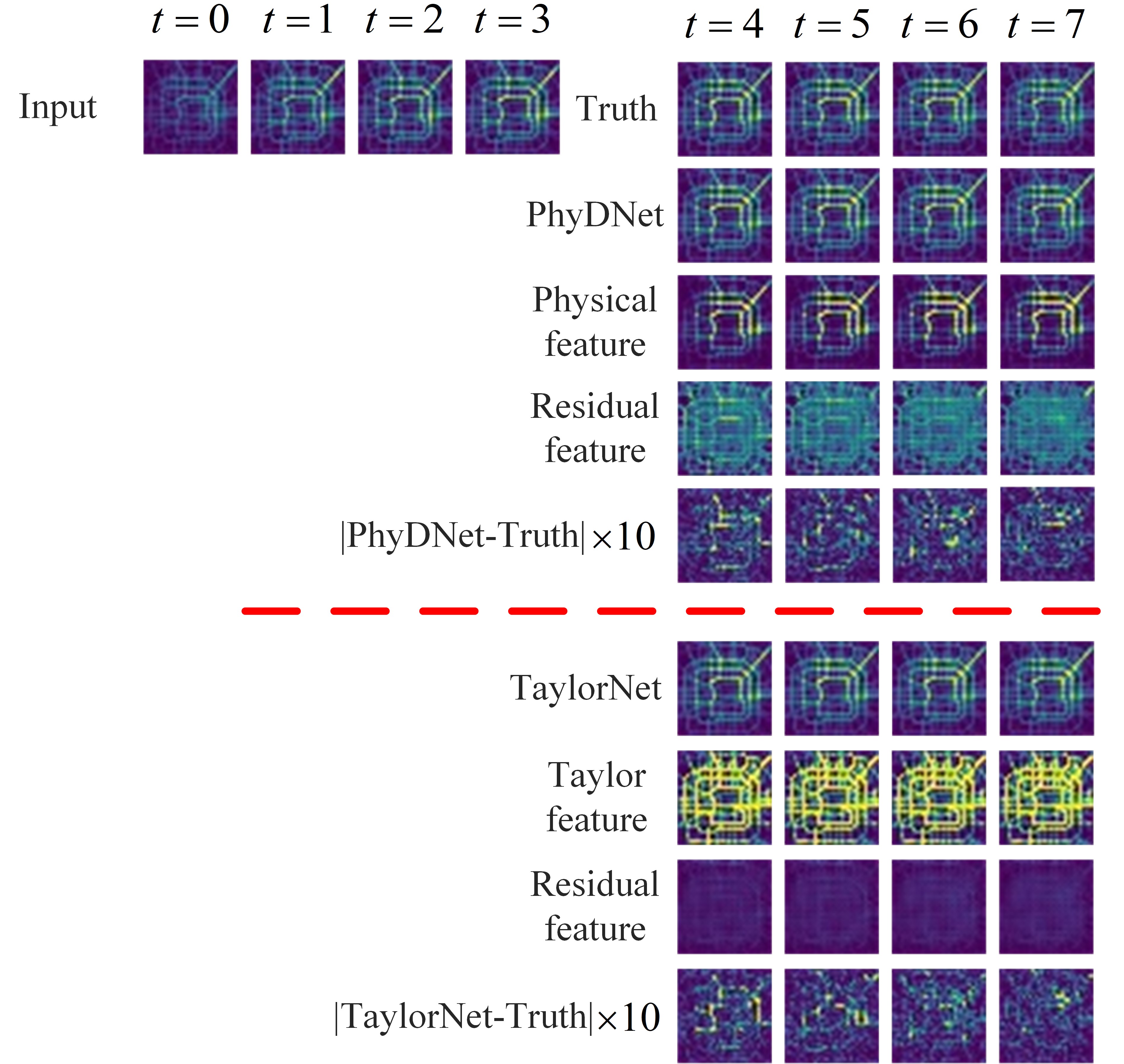}
	\caption{Qualitative comparison on TaxiBJ.}
	\label{fig:taxibj} 
\end{figure}

We compare our model to competitive baselines: (1) ConvLSTM \cite{Shi2015ConvolutionalLN}, (2) PredRNN \cite{Wang2017PredRNNRN}, (3) Causal LSTM \cite{Wang2018PredRNNTA}, (4) MIM \cite{Wang2019MemoryIM}, (5) SA-ConvLSTM\cite{Lange2020AttentionAC}, (6) PDSD \cite{Dona2020PDEDrivenSD}, (7) DDPAE \cite{Hsieh2018LearningTD}, (8) PhyDNet \cite{Guen2020DisentanglingPD}, (9) PhyDNet-dual \cite{Guen2020ADP} and (10) CrevNet \cite{Yu2020EfficientAI}.

Quantitative comparison among different models on Moving MNIST is shown in Table \ref{table:mmnist}.
The optimal results are highlighted in red, whereas the second-optimal results are highlighted in blue.
We can see that TaylorNet outperforms CrevNet, state-of-the-art on Moving MNIST.
Note that CrevNet absence of spatiotemporal LSTM \cite{Wang2017PredRNNRN} is impotent to achieve superior performance, whereas TaylorNet only employs vanilla ConvLSTM.
Quantitative comparison on TaxiBJ is shown in Table \ref{table:taxibj}.
The result of TaylorNet is comparable to the performance of SA-ConvLSTM, state-of-the-art on TaxiBJ.
However, SA-ConvLSTM utilizes high-capacity temporal architectures and has three times more parameters than our model.
Quantitative comparison on Human 3.6 is shown in Table \ref{table:human}.
TaylorNet also reaches very close performance with PhyDNet, which is the state-of-the-art model on Human 3.6.



Moreover, we compare the visualizations of TaylorNet and PhyDNet-dual, which is state-of-the-art in long-term prediction and approaches CrevNet in short-term prediction on Moving MNIST.
Fig \ref{fig:mmnist} demonstrates the qualitative comparison on Moving MNIST.
The third and fourth rows of the figure indicate the visualizations of the Taylor feature and the residual feature, respectively.
The Taylor feature highlights more information prone to ambiguity than the physical feature in PhyDNet-dual. 
The residual feature in our model also extracts more clear content information. 
It proves that the Taylor feature plays a better auxiliary role in modeling the residual feature. 
We can interpret the phenomenon as the difficulty of modeling the residual feature will decrease in the presence of a continuous stability base function of the Taylor feature.
Fig \ref{fig:taxibj} demonstrates the qualitative comparison on TaxiBJ. 
The fifth and ninth rows of the figure indicate the difference $ | $Prediction-Target$ | $$ \times10 $ for better visualization.

\begin{figure}[t]
	\centering
	\includegraphics[width=1.0\linewidth]{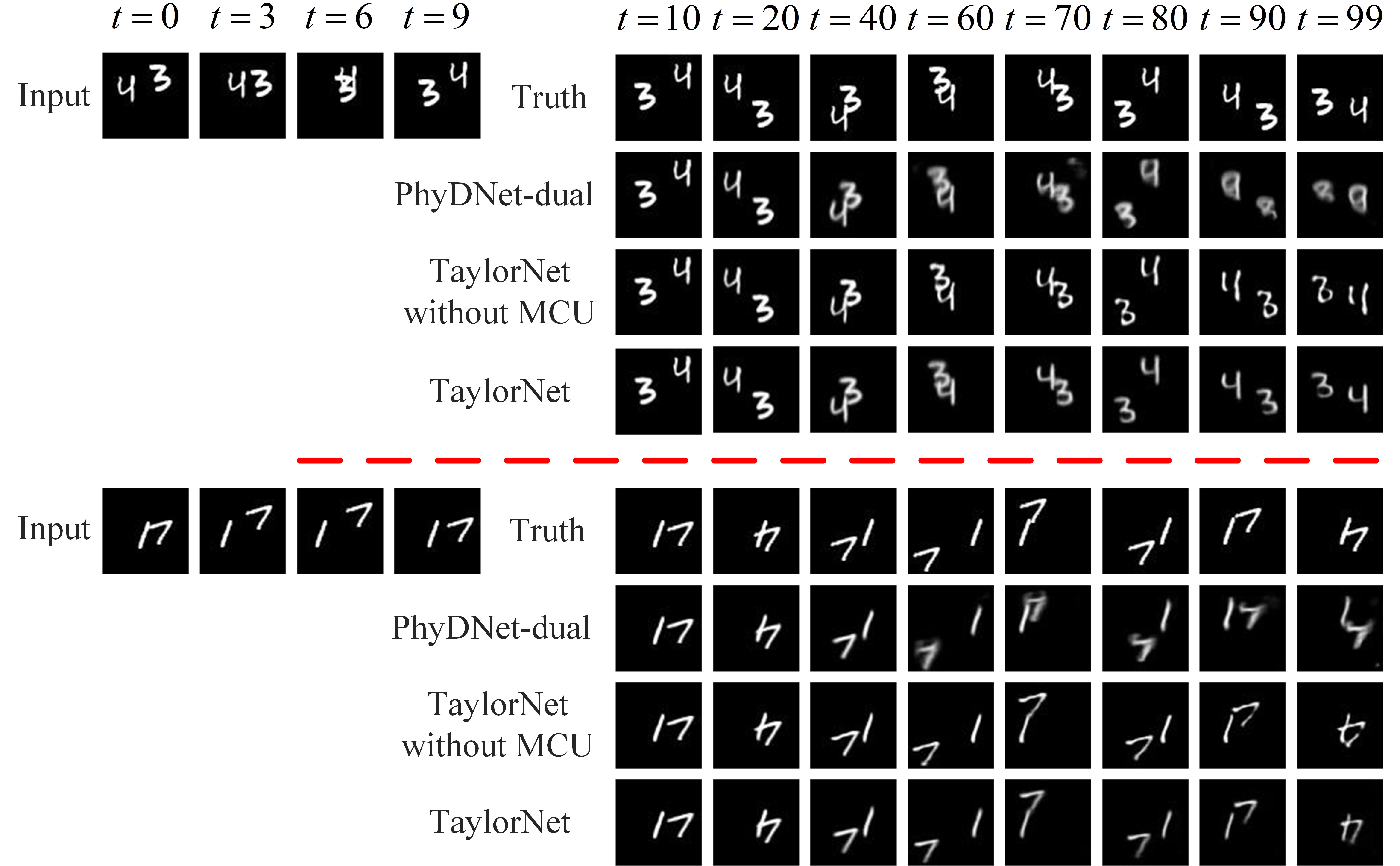}
	\caption{Qualitative comparison in long-term prediction.}
	\label{fig:long-term} 
\end{figure}

\begin{table*}[]
	\caption{Quantitative comparison in long-term prediction. All models predict 10, 30 and 90 frames with 10 conditioning frames.}
	\centering
	\begin{tabular}{c|ccc|ccc|ccc}
		\hline
		\multirow{2}{*}{\textbf{Method}} & \multicolumn{3}{c|}{\textbf{10$ \to $10}} & \multicolumn{3}{c|}{\textbf{10$ \to $30}} & \multicolumn{3}{c}{\textbf{10$ \to $90}} \\ 
		& \textbf{MSE} & \textbf{MAE} & \textbf{SSIM}         & \textbf{MSE} & \textbf{MAE} & \textbf{SSIM}        &\textbf{MSE} & \textbf{MAE} & \textbf{SSIM}        \\ 
		\hline
		PDSD         & 42.9      & 105.6    &  0.878         & 91.8        & 165.3        & 0.792      &179.8         & 268.2       & 0.657       \\
		PhyDNet-dual            & 23.5        & 68.6        & 0.948        & 44.8        & 111.5        & 0.897       & \color{blue}94.3        & 189.5        & 0.797       \\
		TaylorNet without MCU   & \color{blue}22.3        & \color{blue}65.4        & \color{red}0.952        & \color{blue}42.9        & \color{red}100.7        & \color{red}0.909       & 98.8        & \color{red}166.8        & \color{red}0.822       \\
		TaylorNet(ours)         & \color{red}22.2        & \color{red}65.2        & \color{red}0.952        & \color{red}42.2        & \color{blue}105.1        & \color{blue}0.905       & \color{red}91.2        & \color{blue}172.7        & \color{blue}0.812       \\ \hline
	\end{tabular}
	\label{table:MCU}
\end{table*}
\begin{table*}[t]
	\caption{Performance of TaylorCell. All models predict 10 frames with 10 conditioning frames.}
	\centering
	\begin{tabular}{c|c c c c c| c }
		\hline
		\textbf{Method} & \textbf{MSE} & \textbf{MAE} & \textbf{SSIM} & \textbf{BCE} & \textbf{PSNR} & \textbf{PARAMETERS(M)}\\
		\hline
		PhyCell         &  50.8     & 129.3  &  0.87 &-&-& 0.37 \\
		PhyDNet-dual & \color{blue}23.5    &   \color{blue}68.6  &    \color{blue}0.948 & \color{blue}163.6& \color{blue}23.5& 3.09\\
		TaylorCell     & 39.6     & 103.8   &   0.907 & 215.1& 20.9&0.51 \\
		TaylorNet      &  \color{red}22.2      &  \color{red}65.2  &  \color{red}0.952& \color{red}159.5 & \color{red}23.8&3.31  \\
		\hline
	\end{tabular}
	\label{table:TaylorCell}
\end{table*}

\subsection{Ablation Study}
\label{Ablation Study}
We conduct a series of ablation studies on Moving MNIST to show the contribution of several design components in the proposed TaylorNet.

\textbf{Long-term Prediction}	We explore the robustness of TaylorNet in long-term prediction and apply the pre-trained model (10 known frames to predict the next 10 frames) to predict the future 10, 30 or 90 frames, respectively.
As explained in section \ref{MCU}, TaylorNet allows two prediction modes for the Taylor feature: driven by the Taylor inferred information or driven by all past observations.
The first prediction mode overcomes error accumulation because only the first input frame's derivatives and corresponding time parameters are utilized.
The second prediction mode handles unreliable Taylor inferred information due to the limited convergence domain or factors changing.

We compare TaylorNet to PhyDNet-dual, state-of-the-art in long-term prediction.
The second and fourth rows of table \ref{table:MCU} denote the results of TaylorNet and PhyDNet-dual on Moving MNIST.
Compared to PhyDNet-dual, our model gains 1.3 MSE points in predicting 10 frames, 2.6 MSE points in predicting 30 frames and 3.1 MSE points in predicting 90 frames.
TaylorNet also presents consistent gains in MAE and SSIM, indicating the benefits of Taylor series priors.
We also compare TaylorNet to PDSD, which is inspired by PhyDNet. 
TaylorNet outperforms it on all metrics.
Fig \ref{fig:long-term} shows the visualizations of TaylorNet and PhyDNet-dual predicting the next 90 frames.
The quality of output images slightly decreases for TaylorNet, whereas it is much more pronounced for PhyDNet-dual.
For example, PhyDNet for 60-steps has acquired relative blurred predicted frames; however, TaylorNet for 99-steps still generates accurate prediction.

\textbf{Influence of MCU}	To evaluate the contribution of MCU, we delete MCU in TaylorCell and utilize the previous frame instead of all past frames to correct the Taylor inferred feature.
The quantitative results are presented in the third column of Table \ref{table:MCU},  which demonstrates that TaylorNet without MCU gains more MAE and SSIM points in long-term prediction yet increases the MSE.
From the visualizations of Fig \ref{fig:long-term}, we can see that TaylorNet suffers from more prediction uncertainty than TaylorNet without MCU.
In simpler terms, MCU decreases the confidence of prediction in our model, leading to more ambiguity in long-term prediction.
However, TaylorNet without MCU has the crisis of lacking partial image content. 
For example, there are many breakpoints, unexpected connection points, and scabrous edges in predicted handwritten digits, as shown in the third and seventh rows of the Fig \ref{fig:long-term}. 
MCU improves this problem by adding past frames' Taylor information as supplements.
Considering that breakpoints and unexpected connection points may interfere with downstream tasks, MCU is necessary.


\textbf{TaylorCell Analysis}	To more intuitively demonstrate the advantages of our TaylorCell, we conduct an ablation study to analyze the respective performances of TaylorCell and PhyCell.
The results are shown in Table \ref{table:TaylorCell}.
Compared to PhyCell, TaylorCell gains 11.2 MSE points, 25.5 MAE points, and 0.037 SSIM points, while they have a similar number of parameters.
This demonstrates that TaylorCell is a potent recurrent module that outperforms PhyCell due to effective Taylor series priors. 
We further observe the performance gap on TaylorCell and TaylorNet, proving the necessity of the residual module. 
Note that TaylorNet runs faster than PhyDNet-dual by 15 FPS in test.

\begin{figure}[t]
	\centering
	\includegraphics[width=1.0\linewidth]{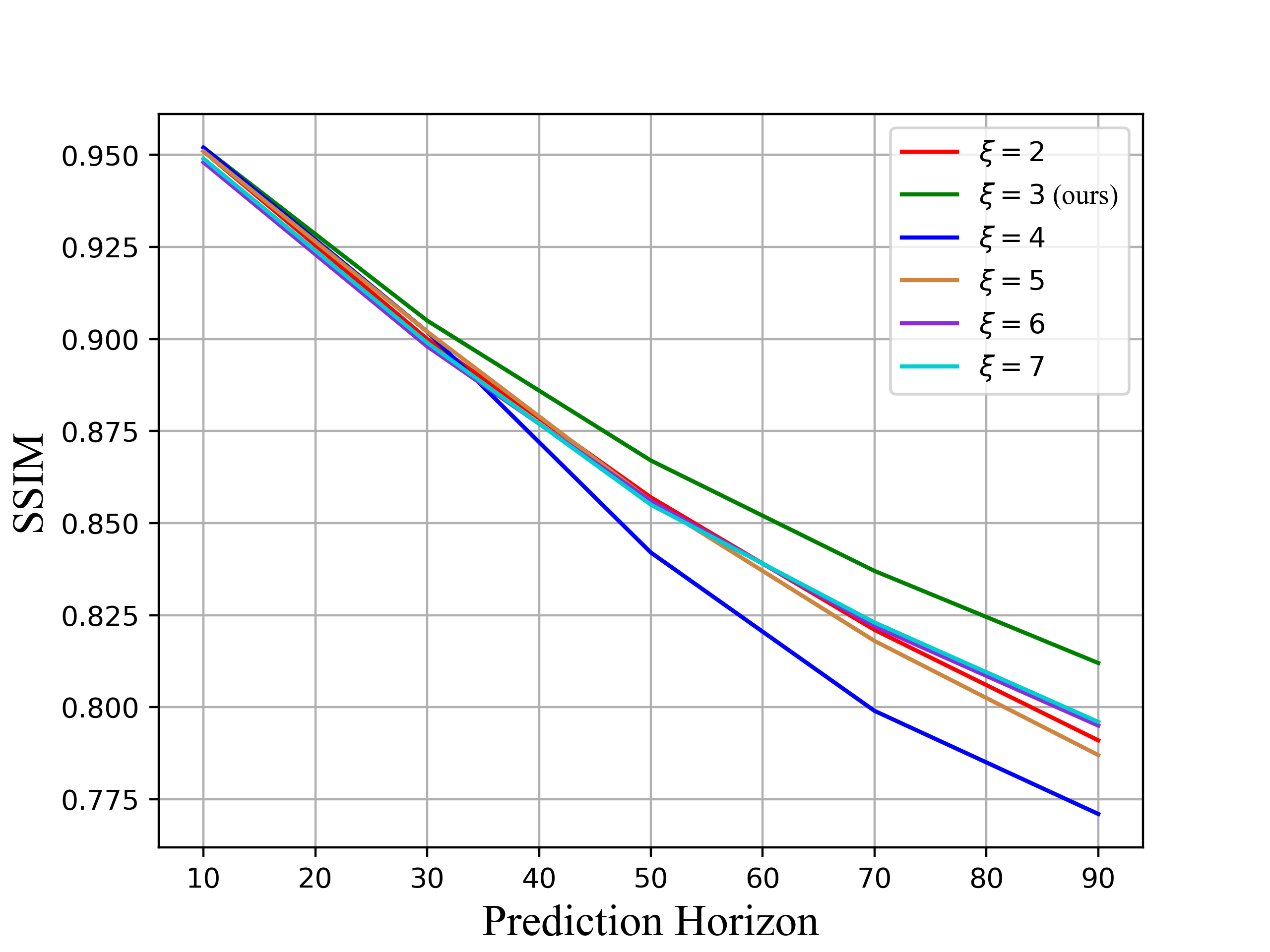}
	\caption{Influence of the expansion order of the Taylor series}
	\label{fig:T_order} 
\end{figure}

\textbf{Orders of the Taylor Series}	Finally, we search the influence of the expansion order of the Taylor series in TPU.
The results are shown in Fig \ref{fig:T_order}, indicating that the more order of Taylor expansions fails to perform better, inconsistent with our intuition.
There exist two reasons: 
(1) the Taylor series is limited to the convergence region and leads to constrained long-term prediction;
(2) too many expansion orders lead to a more significant difference between the Taylor feature and real complex chaotic dynamical systems in long time steps, which meddle with modeling the residual feature.
In fact, the Taylor branch makes the residual component more accessible by modeling the Taylor feature, which is different in each dataset.
For example, our residual branch distills more accurate information about both content and location than that of PhyDNet-dual, as shown in Fig \ref{fig:mmnist}.
Besides, Fig \ref{fig:T_order} also indicates that the performance of TaylorNet is incapable of decreasing step by step, as the expansion order of the Taylor series increases.
However, we fail to explain this phenomenon due to the lack of the interpretability theory of deep neural networks.
Further study is required.
In my shallow opinion, if we plugged a visualization module after the Taylor branch to present each order of the Taylor feature, more explanation might be learned.

\section{Conclusion}
\label{Conclusion}

In this work, we propose a novel two-branch seq-to-seq deep model, TaylorNet, for pixel-level prediction of future frames in videos. 
The originality of our model lies in a novel principle for feature separation and a proposed recurrent prediction module.
Such module enables to model Taylor representations based on Taylor series priors for trade-off short-term and long-term predictions and decreasing error accumulation.
Experimental results demonstrate that TaylorNet outperforms or reaches state-of-the-art performances on three generalist datasets.
Future work includes building distributions over possible futures, \textit{e.g.} with VAE-based or GAN-based features.

\section{Acknowledgements}
\label{Acknowledgements}
Funding: This work was supported by the Ministry of Science and Technology of the People’s Republic of China [grant number 2019YFC1511404]; and the National Natural Science Foundation of China [grant number 62002026].

{\small
	\bibliographystyle{ieee_fullname}
	\bibliography{biblio.bib}
}

\end{document}